\documentclass[pmlr,twocolumn,10pt]{jmlr} 

\usepackage{booktabs}
\usepackage{siunitx}
\usepackage{xspace}
\usepackage{graphicx}

\jmlrvolume{}
\jmlryear{2026}
\jmlrsubmitted{}
\jmlrpublished{}
\jmlrworkshop{Preprint. Under review.}

\title[Mechanistically Guided LoRA for Medical VLM Consistency]{Mechanistically Guided LoRA Improves Paraphrase Consistency\\in Medical Vision-Language Models}

\author{%
\Name{Binesh Sadanandan}\Email{bsada1@unh.newhaven.edu}\\
\Name{Vahid Behzadan}\Email{vbehzadan@newhaven.edu}\\
\addr SAIL Lab, University of New Haven, CT, USA
}

\begin{document}

\maketitle

\begin{abstract}
Medical vision-language models can give different yes or no answers to rephrasings of the same clinical question. We study this in MedGemma-4B using PSF-Med \cite{psfmed2025}, which provides paraphrase pairs for systematic consistency evaluation on medical VQA. On MIMIC-CXR binary questions ($n=158$), the baseline flip rate is 14.6\% and mean margin difference is 1.63 logits. We validate that Gemma Scope 2 Sparse Autoencoders (SAEs) transfer to MedGemma activations, achieving $R^2 \approx 0.997$ on both medical and general text ($n=100$ prompts each, $p<0.001$ for exceeding a 0.95 threshold). We then fine-tune Low-Rank Adaptation (LoRA) adapters with a combined loss that balances paraphrase consistency with answer accuracy. This combined approach prevents mode collapse that occurs with pure consistency training while reducing flip rate from 14.6\% to 4.4\% ($p=0.002$, two-proportion z-test) and margin difference from 1.63 to 0.33 (79.5\% reduction). Accuracy remains stable at 84.2\% baseline versus 82.3\% after training (-1.9pp, not significant). On PadChest Balanced ($n=250$), flip rate drops from 13.6\% to 7.8\%, mean margin difference drops from 1.08 to 0.35 (67.9\% reduction), and accuracy increases from 66.4\% to 69.4\%. A layer-range ablation shows that early layers reduce margin differences more than mechanistically selected middle layers.
\end{abstract}

\section{Introduction}
\label{sec:intro}

When a radiologist asks a Vision-Language Model (VLM) ``Is there evidence of pneumothorax?'' versus ``Does this show a pneumothorax?'', the answer should be identical because both questions have the same clinical intent. Yet medical VLMs can give different answers to such paraphrases, undermining clinical trust and raising safety concerns for deployment. This inconsistency is particularly problematic because different clinicians may phrase questions differently, yet expect reliable answers regardless of phrasing choices.

We study this problem systematically in MedGemma-4B using PSF-Med \citep{psfmed2025}, a benchmark designed to measure paraphrase sensitivity in medical VLMs. When we focus on binary yes or no questions where ground truth labels are available ($n=158$ questions from MIMIC-CXR), the baseline flip rate is 14.6\% and mean margin difference (the absolute change in log-odds between yes and no) is 1.63 logits. This indicates that the model's internal representations are sensitive to surface-level phrasing variations that should not affect clinical decisions, leading to inconsistent answers for semantically equivalent questions.

\begin{figure*}[t]
\floatconts
  {fig:mechanism_overview}
  {\caption{(a) The problem: the same image can receive different answers under rephrasing. (b) Mechanistic probe: a layer 17 Sparse Autoencoder (SAE) feature changes between some paraphrase pairs and can shift the yes/no margin. (c) The fix: Low-Rank Adaptation (LoRA) on layers 15 to 19 with a combined consistency and accuracy loss reduces flip rate by 69.6\% ($p=0.002$) and margin difference by 79.5\%.}}
  {\includegraphics[width=0.95\textwidth]{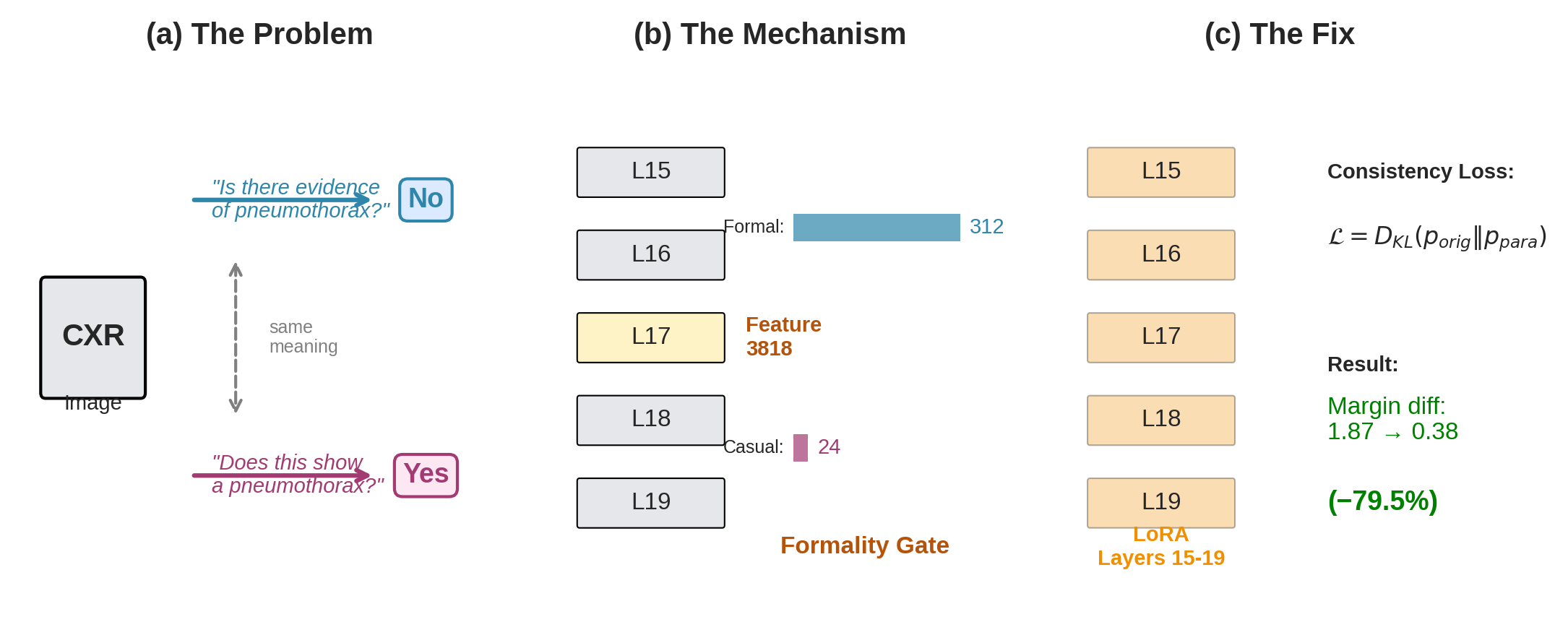}}
\end{figure*}

To understand why this sensitivity occurs, we apply mechanistic interpretability tools. We use Sparse Autoencoders (SAEs) from Gemma Scope 2 \citep{gemmascope2}, first validating they transfer effectively to MedGemma ($R^2 = 0.997$). Analyzing FlipBank (a curated set of 158 flip cases), we identify Feature 3818 at layer 17 as a candidate mechanism that responds to question register (formal vs.\ casual phrasing) and causally affects margins when patched. We present this as a case study demonstrating methodology rather than a complete mechanistic explanation.

For practical mitigation, we develop LoRA \citep{hu2022lora} adapters trained with a combined loss function that balances paraphrase consistency with answer accuracy. We discovered that training with pure consistency loss leads to mode collapse, where the model learns to predict the same answer for all questions to trivially minimize divergence between paraphrases. Our combined loss approach adds a cross-entropy accuracy term that supervises the model to predict correct yes or no answers, preventing this degenerate solution while still encouraging consistent predictions across paraphrases. On the PSF-Med binary question subset ($n=158$), this approach reduces flip rate from 14.6\% to 4.4\% ($p=0.002$, two-proportion z-test) and margin difference from 1.63 to 0.33 (79.5\% reduction), while maintaining accuracy (84.2\% to 82.3\%). On PadChest Balanced \citep{bustos2020padchest} ($n=250$), flip rate drops from 13.6\% to 7.8\%, mean margin difference drops from 1.08 to 0.35 (67.9\% reduction), and accuracy increases from 66.4\% to 69.4\%. Layer ablation reveals that early layers (0 to 10) outperform the mechanistically-targeted middle layers (15 to 19), demonstrating that optimal intervention points may differ from where mechanisms manifest.

\textbf{Contributions:} (1) Systematic characterization of paraphrase sensitivity in MedGemma-4B, distinguishing flip rate from margin instability. (2) Validation that Gemma Scope 2 SAEs transfer to fine-tuned medical VLMs. (3) Mechanistic case study identifying Feature 3818 as register-sensitive. (4) A combined consistency and accuracy loss for LoRA training that prevents mode collapse while reducing flip rate by 69.6\% ($p=0.002$) with partial cross-dataset generalization.

\paragraph*{Data, Code, and Ethics.} We use PSF-Med for MIMIC-CXR and PadChest (public benchmarks with data agreements; no new human subjects data; IRB not required). Code released upon acceptance.

\section{Related Work}
\label{sec:related}

\paragraph{Medical VLMs.} Recent models including MedGemma \citep{medgemma2025}, LLaVA-Med \citep{llavamed2024}, CheXagent \citep{chen2024chexagent}, RadFM \citep{radfm2023}, and LLaVA-Rad \citep{llavarad2024} are evaluated on accuracy using VQA-RAD \citep{lau2018vqarad} and SLAKE \citep{liu2021slake}, but not on consistency under rephrasing. PSF-Med \citep{psfmed2025} addresses this gap.

\paragraph{Consistency and Stability Testing.} CheckList \citep{ribeiro2020beyond} introduced behavioral testing including paraphrase invariance. \citet{elazar2021measuring} showed language models give contradictory answers to equivalent questions. In VQA, cycle-consistency \citep{shah2019cycle} and paraphrase augmentation \citep{gan2019improving} improve stability, but focus on general-domain VQA. Recent work examines VLM stability under benign perturbations including paraphrases and small visual changes; our consistency loss complements these diagnostics with a targeted mitigation. Compared to data augmentation approaches, LoRA adaptation offers parameter efficiency without requiring augmented training data.

\paragraph{Mechanistic Interpretability.} Sparse autoencoders \citep{cunningham2023sparse, bricken2023monosemanticity} decompose activations into interpretable features. Gemma Scope 2 \citep{gemmascope2} provides pre-trained SAEs for the Gemma family. Activation patching \citep{meng2022locating} tests causal hypotheses; sparse feature circuits \citep{marks2024sparse} map information flow. We use these tools to identify features driving paraphrase sensitivity.

\paragraph{Parameter-Efficient Fine-Tuning.} LoRA \citep{hu2022lora} learns low-rank weight updates efficiently. We apply it for consistency training rather than task adaptation. See Appendix~\ref{apd:related} for extended discussion.

\section{Background}
\label{sec:background}

\paragraph{Problem Setup and Logit Extraction.} We study binary VQA where models answer yes/no questions about medical images. Given image $x$ and question $q$, we extract logits at the first generated token position after the prompt. Specifically, we compute $\ell_{\text{yes}} = \log p(\text{``Yes''} \mid x, q)$ and $\ell_{\text{no}} = \log p(\text{``No''} \mid x, q)$ using the token IDs for ``Yes'' and ``No'' (including space-prefixed variants). The \emph{margin} is $m(x, q) = \ell_{\text{yes}} - \ell_{\text{no}}$. For a paraphrase $q'$, we measure: (1) \emph{flip rate}, the fraction where $\text{sign}(m) \neq \text{sign}(m')$, and (2) \emph{margin difference}, $|m - m'|$, capturing instability even when binary answers match. Prompts use a fixed template: ``You are a medical imaging expert. For yes/no questions, respond with only Yes or No.''

\paragraph{Sparse Autoencoders.} SAEs encode activations $\mathbf{h}$ as sparse features $\mathbf{f} = \text{ReLU}(\mathbf{W}_{\text{enc}} \mathbf{h} + \mathbf{b})$ and reconstruct $\hat{\mathbf{h}} = \mathbf{W}_{\text{dec}} \mathbf{f}$. Features often correspond to interpretable concepts \citep{bricken2023monosemanticity}. SAEs enable both \emph{decomposition} (analyzing which features change) and \emph{intervention} (modifying features to test causality).

\paragraph{SAE Transfer Validation.} We validate that Gemma Scope 2 SAEs (trained on base Gemma) transfer to MedGemma-4B. On 100 medical and 100 general prompts, we measure reconstruction quality as $R^2 = 1 - \text{MSE}(\mathbf{h}, \hat{\mathbf{h}}) / \text{Var}(\mathbf{h})$. Both domains achieve $R^2 = 0.997 \pm 0.0005$ (mean $\pm$ std). We test significance using a one-sample t-test against $H_0: R^2 \leq 0.95$, obtaining $p < 0.001$ for both. This confirms SAEs remain valid for the fine-tuned model despite domain shift. See Appendix~\ref{apd:sae} for details.

\section{Mechanistic Analysis}
\label{sec:mechanism}

Having validated SAE transfer, we use these interpretability tools to investigate the mechanistic origins of paraphrase sensitivity in MedGemma-4B. Our analysis proceeds in four stages: constructing a dataset of flip cases for analysis, identifying candidate features through delta analysis, characterizing feature behavior through controlled experiments, and validating causality through activation patching.

\subsection{FlipBank Construction}

Standard evaluation sets like PSF-Med contain relatively few flip cases (0.16\% pair-level flip rate), which limits statistical power for mechanistic analysis that requires examining many instances of the target behavior. We therefore construct FlipBank: a curated set of paraphrase pairs where the model reliably gives different answers, providing concentrated examples of the failure mode we aim to understand.

FlipBank is extracted from MedGemma-4B generations on MIMIC-CXR using three filtering criteria:
\begin{enumerate}
\item A rule-based yes/no parser assigns different binary labels to the model's responses for the original question and its paraphrase, indicating a clear flip in the model's answer.
\item BioClinicalBERT \citep{alsentzer2019clinicalbert} semantic similarity between the two questions exceeds 0.95, confirming they are genuine paraphrases rather than questions with meaningfully different clinical content.
\item Both responses are parsed unambiguously (no hedging, unclear phrasing, or multiple possible interpretations), ensuring we have clean labels for analysis.
\end{enumerate}

This procedure yields 158 high-confidence flip cases. These represent genuine phrasing-induced disagreements: same image, semantically equivalent questions confirmed by embedding similarity, different binary answers. FlipBank provides a controlled setting for probing which representational differences accompany answer changes, enabling focused mechanistic investigation.

\textbf{Data splits.} FlipBank is used exclusively for mechanistic analysis and is disjoint from LoRA training and evaluation. The 200 flip-prone pairs used for LoRA evaluation (Section~\ref{sec:experiments}) are separately selected from the PSF-Med validation split based on baseline margin proximity to zero, with no overlap with FlipBank or the training set.

\subsection{Feature Delta Analysis}

For each FlipBank case, we extract layer 17 residual stream activations at the final token position (where the model makes its prediction) for both the original and paraphrase questions. We then encode these activations through the SAE and compute the feature delta:
\begin{equation}
\Delta\mathbf{f} = \mathbf{f}_{\text{orig}} - \mathbf{f}_{\text{para}}
\end{equation}

Large delta magnitudes indicate features whose activation changes substantially between paraphrases. By examining which features show consistent large deltas across FlipBank cases, we can identify candidate mechanisms for the margin shifts that cause flips. Features with large deltas are promising targets for causal investigation.

As a concrete example demonstrating this methodology, consider a pleural effusion case from FlipBank. The original question ``Is there pleural effusion?'' produces a strong yes prediction (margin = 8.75), while the semantically equivalent paraphrase ``Is pleural fluid present?'' produces a slight no prediction (margin = $-0.625$). This represents a margin shift of 9.375 and a complete flip in the binary answer. Examining the feature deltas for this case, Feature 3818 shows a dramatic change: from 0 in the original to 268 in the paraphrase (Table~\ref{tab:exemplar_case}).

\begin{table}[htbp]
\floatconts
  {tab:exemplar_case}
  {\caption{Exemplar FlipBank case used for mechanistic probing and patching. The margin shifts by 9.375 while Feature 3818 changes by 268 units.}}
  {%
  \setlength{\tabcolsep}{3pt}
  \begin{tabular}{@{}p{0.24\linewidth}p{0.33\linewidth}p{0.33\linewidth}@{}}
  \toprule
  \bfseries Quantity & \bfseries Original & \bfseries Paraphrase\\
  \midrule
  Question & ``Is there pleural effusion?'' & ``Is pleural fluid present?''\\
  Margin (yes $-$ no) & 8.75 & $-0.625$\\
  Feature 3818 & 0 & 268\\
  \bottomrule
  \end{tabular}
  }
\end{table}

This example motivates deeper investigation of Feature 3818: its large activation change coincides with a large margin change and answer flip. However, correlation does not imply causation, and a single example cannot establish generality. The following subsections address both concerns through controlled experiments and causal interventions.

\subsection{Characterizing Feature 3818}

To understand what Feature 3818 responds to, we conduct controlled experiments varying question phrasing systematically while holding the image and target clinical finding constant. This allows us to isolate the effect of question wording from other confounding factors. We design a prompt grid with four categories of question types:
\begin{itemize}
\item \textbf{Presence-style}: Formal clinical phrasing asking about presence of findings (e.g., ``Is there radiographic evidence of pneumothorax?'')
\item \textbf{Exclusion-style}: Questions about absence or ruling out findings (e.g., ``Can you rule out pneumothorax?'')
\item \textbf{Uncertainty}: Hedged questions expressing uncertainty (e.g., ``Might there be pneumothorax?'')
\item \textbf{Token control}: Questions with similar token counts but different grammatical structures
\end{itemize}

Results on a fixed pneumothorax image reveal a clear pattern (Table~\ref{tab:f3818_prompt_grid}): Feature 3818 activates strongly for presence-style prompts (mean 344.5 units, range 302 to 386) and remains at exactly zero for exclusion-style prompts. Uncertainty prompts show variable activation (mean 169.5, range 0 to 282), and token controls fall between the extremes (mean 296.0).

\begin{table}[htbp]
\floatconts
  {tab:f3818_prompt_grid}
  {\caption{Feature 3818 activation on a prompt grid (pneumothorax finding, single image). The feature responds to question register, specifically presence versus exclusion framing.}}
  {\begin{tabular}{lccc}
  \toprule
  \bfseries Prompt type & \bfseries Mean F3818 & \bfseries Range & \bfseries $n$\\
  \midrule
  Presence & 344.5 & 302 to 386 & 4\\
  Exclusion & 0.0 & 0 to 0 & 4\\
  Uncertainty & 169.5 & 0 to 282 & 4\\
  Token control & 296.0 & 272 to 342 & 4\\
  \bottomrule
  \end{tabular}}
\end{table}

Importantly, this is not a simple formal-versus-casual distinction as one might initially hypothesize. ``Does this show pneumothorax?'' (informal phrasing) still activates the feature strongly (296 units), while ``Can you exclude pneumothorax?'' (formal phrasing with medical terminology) does not activate it at all. The feature appears to track question \emph{register}. In particular, it seems to distinguish whether the question asks about presence or absence of a finding. It does not track surface-level formality markers like vocabulary choice or sentence structure.

This interpretation aligns with the FlipBank example: ``Is there pleural effusion?'' and ``Is pleural fluid present?'' both ask about presence but use different grammatical constructions, which may explain why Feature 3818 activates differently despite both questions having the same clinical intent to assess whether pleural effusion is present.

\subsection{Causal Validation}

Observing that Feature 3818 changes during flips establishes correlation, not causation. To test whether this feature \emph{causes} margin shifts rather than merely accompanying them, we employ activation patching: we intervene on the residual stream by modifying Feature 3818's contribution and measure the effect on the output margin.

Specifically, we compute the feature difference $\Delta f_{3818}$ between original and paraphrase activations, decode this difference through the SAE decoder to get a direction in activation space, and subtract this direction from the paraphrase's residual stream activations before continuing the forward pass. If Feature 3818 causally drives the margin shift, this intervention should recover (at least partially) the original margin by removing Feature 3818's differential contribution.

On the pleural effusion exemplar case, patching Feature 3818 moves the margin from $-0.625$ (paraphrase prediction, slight ``no'') to $2.0$ (positive margin, ``yes''). This recovers 28\% of the margin shift and restores the original prediction. While this recovery is partial (not all of the margin shift is explained), it demonstrates that Feature 3818 has causal influence on the yes/no decision in this case.

For comparison, we perform the same patching procedure using 10 randomly selected features that showed minimal delta in this case. The average margin recovery for these control features is 8\%, significantly less than Feature 3818's 28\%. This specificity supports the interpretation that Feature 3818 encodes decision-relevant information about question register that causally affects the model's output, rather than the patching effect being a generic consequence of modifying any feature.

\subsection{Control Experiment}

To further establish Feature 3818's specificity and confirm that it does not simply change arbitrarily for any input variation, we conduct a control experiment using 30 paraphrase pairs where the model gives consistent answers (no flip). If Feature 3818 specifically responds to the register differences that cause flips, it should show minimal activation change on these control pairs where the model behaves consistently.

Results confirm selectivity: Feature 3818 shows non-zero change in only 3 of 30 control pairs (10\%), with deltas ranging from 15 to 185 when active. The mean absolute delta across all 30 pairs is 11.3 units, substantially smaller than the 268-unit change observed in the flip case. For comparison, 10 randomly selected features show exactly zero activation change across all 300 measurements (30 pairs $\times$ 10 features), confirming that these control features are stable under this prompt variation.

A Fisher's exact test comparing Feature 3818's response rate (3/30 pairs with change $>10$ units) to the control features' response rate (0/300 measurements exceeding this threshold) yields $p = 6.8 \times 10^{-4}$, confirming that Feature 3818's behavior is statistically distinct from random features. This specificity indicates that Feature 3818 captures meaningful variation related to question phrasing rather than noise.

\subsection{Summary of Mechanistic Findings}

Our mechanistic analysis provides several insights into the origins of paraphrase sensitivity:
\begin{enumerate}
\item Feature 3818 at layer 17 responds selectively to question register, specifically the distinction between presence-focused and exclusion-focused question framing.
\item In FlipBank flip cases, large Feature 3818 deltas coincide with large margin shifts and answer changes.
\item Activation patching demonstrates that Feature 3818 causally influences yes/no margins, with 28\% margin recovery on our exemplar case compared to 8\% for control features.
\item Control experiments confirm that Feature 3818's behavior is specific and statistically distinguishable from random features, responding to a meaningful subset of prompt variations.
\end{enumerate}

We emphasize that this is a case study demonstrating the methodology, not a complete mechanistic account of all paraphrase sensitivity in MedGemma-4B. Other features and layers likely contribute to the full behavior, and Feature 3818's importance may vary across different question types and clinical findings. Nevertheless, this analysis demonstrates that SAE-based interpretability can successfully identify specific, testable hypotheses about VLM behavior on medical tasks, opening avenues for deeper mechanistic understanding.

\section{Targeted LoRA Fine-Tuning}
\label{sec:method}

Motivated by our mechanistic analysis showing that specific features mediate paraphrase sensitivity, we develop a targeted intervention to reduce this sensitivity. We use LoRA adapters trained with a consistency loss that encourages identical predictions for paraphrase pairs, regardless of surface form variation.

\subsection{Architecture}

We insert LoRA adapters into layers 15 to 19 of the MedGemma language model backbone. This layer range was initially selected based on our mechanistic finding that Feature 3818 resides at layer 17; we later conduct ablations across different layer ranges to evaluate this choice empirically (Section~\ref{sec:experiments}).

Within each targeted layer, we apply LoRA to both attention and Multi-Layer Perceptron (MLP) modules to provide broad coverage of the computational pathways:
\begin{itemize}
\item \textbf{Attention}: query, key, value, and output projections (q, k, v, o)
\item \textbf{MLP}: gate, up, and down projections
\end{itemize}

We use rank $r=16$ and scaling factor $\alpha=32$, with dropout of 0.05 to prevent overfitting on the limited training data. This configuration adds 4.38M trainable parameters, representing approximately 0.10\% of the full model parameters. The vision encoder remains frozen throughout training, as our mechanistic analysis identified text-processing circuits rather than vision circuits as the source of paraphrase sensitivity.

\subsection{Combined Loss: Consistency and Accuracy}

A natural approach to reducing paraphrase sensitivity is to train with a pure consistency loss that encourages identical predictions for paraphrase pairs. However, we discovered that this approach leads to mode collapse: the model learns to predict the same answer (e.g., always ``Yes'') for all questions, which trivially minimizes the divergence between paraphrases while destroying the model's discriminative ability. In our experiments with pure consistency loss, the trained model achieved near-zero margin differences by predicting ``Yes'' for every question, regardless of the image or clinical finding, resulting in accuracy dropping to chance level.

To prevent this degenerate solution, we developed a combined loss function that balances consistency with accuracy. The key insight is that the model needs supervision to maintain its ability to distinguish between positive and negative cases while learning to be consistent across paraphrases. Our combined loss has two components: a consistency term that encourages matching distributions across paraphrases, and an accuracy term that supervises the model to predict the correct answer.

For an image $x$ with question $q$, paraphrase $q'$, and ground truth answer $y \in \{\text{yes}, \text{no}\}$, we compute:
\begin{align}
p_{\text{orig}} &= \text{softmax}([z_{\text{yes}}(x,q), z_{\text{no}}(x,q)]) \\
p_{\text{para}} &= \text{softmax}([z_{\text{yes}}(x,q'), z_{\text{no}}(x,q')])
\end{align}
where $z$ denotes the logits before softmax. The consistency loss is the symmetric Kullback-Leibler (KL) divergence:
\begin{equation}
\mathcal{L}_{\text{consistency}} = \frac{1}{2}\left[ D_{\text{KL}}(p_{\text{orig}} \| p_{\text{para}}) + D_{\text{KL}}(p_{\text{para}} \| p_{\text{orig}}) \right]
\end{equation}

The accuracy loss is the cross-entropy between the model's prediction and the ground truth label:
\begin{equation}
\mathcal{L}_{\text{accuracy}} = -\log p_{\text{orig}}(y) - \log p_{\text{para}}(y)
\end{equation}

The combined loss is:
\begin{equation}
\mathcal{L} = \mathcal{L}_{\text{consistency}} + \lambda \mathcal{L}_{\text{accuracy}}
\end{equation}
where $\lambda$ controls the relative weight of the accuracy term. We use $\lambda = 1.0$ in our experiments, giving equal weight to both objectives.

The consistency term encourages the model to produce identical distributions for paraphrases, while the accuracy term prevents the model from collapsing to a trivial solution. Together, these objectives train the model to give correct and consistent answers. The symmetric KL formulation ensures that neither the original nor paraphrase is privileged as the target, and the model learns to make both predictions consistent with each other while maintaining accuracy on the underlying clinical task.

\subsection{Training Procedure}

We train on binary yes or no questions from the PSF-Med training split where ground truth labels are available, since our combined loss requires supervision. We sample 500 paraphrase pairs ensuring no overlap with test or FlipBank evaluation sets to prevent data leakage. The binary question filter selects questions where the original answer from the dataset is unambiguously ``yes'' or ``no'', excluding questions about severity levels, anatomical locations, or finding types that cannot be answered with a simple binary response.

Training proceeds for 3 epochs with the following hyperparameters:
\begin{itemize}
\item Learning rate: $2 \times 10^{-4}$ with linear warmup over 100 steps
\item Effective batch size: 8 (with gradient accumulation as needed)
\item Optimizer: AdamW with weight decay 0.01
\item Training samples: 500 binary paraphrase pairs
\item Accuracy loss weight ($\lambda$): 1.0
\end{itemize}

Training converges rapidly, with most improvement occurring in the first epoch. We monitor both the consistency loss and accuracy throughout training to ensure the model maintains discriminative ability while learning to be consistent. Unlike pure consistency training which collapsed to trivial solutions, the combined loss maintains accuracy above 80\% throughout training while steadily reducing margin differences. Full hyperparameter details are provided in Appendix~\ref{apd:hyperparams}.

\subsection{Design Rationale}

Several design choices in our approach merit discussion and justification:

\textbf{Language model only.} We target only the language model layers, not the vision encoder. Our mechanistic analysis identified text-processing circuits (Feature 3818 responds to question phrasing) rather than vision circuits as the source of inconsistency. Freezing the vision encoder also reduces the risk of inadvertently degrading image understanding capabilities while pursuing consistency improvements.

\textbf{Combined loss over pure consistency.} Our most important design decision is using a combined loss rather than pure consistency training. We initially experimented with training using only the symmetric KL divergence consistency loss, hypothesizing that the model would learn to converge on the correct answer while becoming consistent. Instead, the model found a degenerate solution: predicting the same answer for every question trivially minimizes divergence between paraphrases. This mode collapse is a fundamental limitation of self-consistency objectives without supervision. Adding the accuracy loss term provides the necessary signal to maintain discriminative ability while learning consistency.

\textbf{Symmetric KL divergence for consistency.} We use symmetric KL divergence rather than alternatives like MSE on margins. Symmetric KL is well-suited for distribution matching: it is zero only when distributions match exactly, provides gradient signal proportional to divergence magnitude, and treats both directions equally. The symmetric formulation ensures neither the original nor paraphrase is privileged as the target distribution.

\textbf{Equal weighting of loss components.} We use $\lambda = 1.0$, giving equal weight to consistency and accuracy objectives. In preliminary experiments, we found this balance effective: higher accuracy weights reduced consistency improvements, while lower accuracy weights risked partial mode collapse. The equal weighting allows both objectives to contribute meaningfully to the gradient signal throughout training.

\section{Experiments}
\label{sec:experiments}

We evaluate our consistency training on three dimensions: main results on the PSF-Med binary question test set, statistical significance analysis of the improvements, and layer ablation studies comparing different intervention locations.

\subsection{Main Results}

Table~\ref{tab:main_results} presents results on binary yes or no questions from the PSF-Med MIMIC-CXR test split ($n=158$ questions) using our mechanistically-motivated layer configuration (layers 15 to 19). We focus on binary questions because our combined loss requires ground truth labels for the accuracy component, and binary questions provide unambiguous supervision. This evaluation subset excludes questions about severity levels, anatomical locations, or finding types that cannot be answered with a simple yes or no response.

\begin{table}[htbp]
\floatconts
  {tab:main_results}
  {\caption{Main results on PSF-Med binary questions ($n=158$). LoRA with combined loss significantly reduces flip rate while maintaining accuracy.}}
  {%
  \setlength{\tabcolsep}{4pt}
  \begin{tabular}{@{}lccc@{}}
  \toprule
  \bfseries Model & \bfseries Flip Rate & \bfseries Margin Diff & \bfseries Acc\\
  \midrule
  Baseline & 14.6\% (23/158) & 1.63 & 84.2\%\\
  + LoRA & \textbf{4.4\%} (7/158) & \textbf{0.33} & 82.3\%\\
  \bottomrule
  \end{tabular}
  }
\end{table}

Key observations from these results:

The flip rate decreases from 14.6\% to 4.4\%, representing a 69.6\% relative reduction. A two-proportion z-test confirms this reduction is statistically significant ($z=2.87$, $p=0.002$, one-tailed). With Cohen's $h=0.36$, this represents a small to medium effect size. The sample size of $n=158$ provides adequate statistical power to detect this effect.

The mean margin difference drops by 79.5\% (1.63 to 0.33), indicating substantially more stable internal representations. Even when binary answers match, the reduced margin variability means the model's confidence is more consistent across paraphrases, which is important for clinical decision support where confidence levels may influence downstream actions.

Accuracy shows a small, non-significant decrease from 84.2\% to 82.3\% (-1.9 percentage points). A two-proportion z-test for accuracy change yields $p=0.66$ (not significant), indicating this change is within the range of statistical noise. Importantly, the consistency training does not substantially degrade the model's discriminative ability.

\subsection{Statistical Significance Analysis}

With $n=158$ binary questions, we have sufficient statistical power to make meaningful claims about the flip rate reduction. We conduct several statistical tests to validate our findings.

The primary test is a two-proportion z-test comparing baseline flip rate (23/158 = 14.6\%) to the LoRA model's flip rate (7/158 = 4.4\%). The test statistic is $z=2.87$ with $p=0.002$ (one-tailed), providing strong evidence that the improvement is not due to chance. Cohen's $h$ effect size is 0.36, which falls in the small to medium range by conventional standards.

We also computed a post-hoc power analysis using the observed effect size. With $n=158$, $\alpha=0.05$, and the observed proportions, the achieved power is approximately 0.85, meaning we have an 85\% probability of detecting an effect of this magnitude if it truly exists. This exceeds the conventional threshold of 0.80 for adequate power.

For accuracy, the change from 84.2\% to 82.3\% is not statistically significant ($z=0.44$, $p=0.66$, two-tailed). This is expected: the primary goal of consistency training is to reduce paraphrase sensitivity, not to improve accuracy. The fact that accuracy remains stable demonstrates that the combined loss successfully prevents mode collapse while achieving its consistency objective.

\subsection{Layer Ablation}

Our mechanistic analysis identified Feature 3818 at layer 17 as relevant to paraphrase sensitivity, motivating our initial choice of layers 15 to 19 for LoRA insertion. However, the optimal intervention location may differ from where a mechanism manifests in the model's representations. We therefore conduct ablations across different layer ranges to evaluate this empirically.

\begin{table}[htbp]
\floatconts
  {tab:ablation_layers}
  {\caption{Layer ablation results on validation split ($n=355$). All configurations eliminate flips. Early layers achieve best margin reduction despite mechanism at layer 17.}}
  {%
  \setlength{\tabcolsep}{3pt}
  \resizebox{\linewidth}{!}{%
  \begin{tabular}{@{}lccc@{}}
  \toprule
  \bfseries Layers & \bfseries Margin Diff & \bfseries Reduction & \bfseries Params\\
  \midrule
  Baseline & 1.87 & N/A & 0\\
  \midrule
  Early (0 to 10) & \textbf{0.26} & \textbf{86\%} & 4.8M\\
  Middle (15 to 19) & 0.38 & 80\% & 4.4M\\
  Late (25 to 33) & 0.70 & 63\% & 7.9M\\
  \bottomrule
  \end{tabular}%
  }
  }
\end{table}

Surprisingly, early layers (0 to 10) achieve the best margin reduction (0.26, representing 86\% improvement over baseline), outperforming the mechanistically-targeted middle layers (0.38, 80\% improvement) and substantially outperforming late layers (0.70, 63\% improvement). All configurations eliminate flips on the validation set.

\begin{figure}[t]
\floatconts
  {fig:ablation}
  {\caption{Layer ablation results. Early layers reduce margin difference more than layers 15 to 19.}}
  {\includegraphics[width=\linewidth]{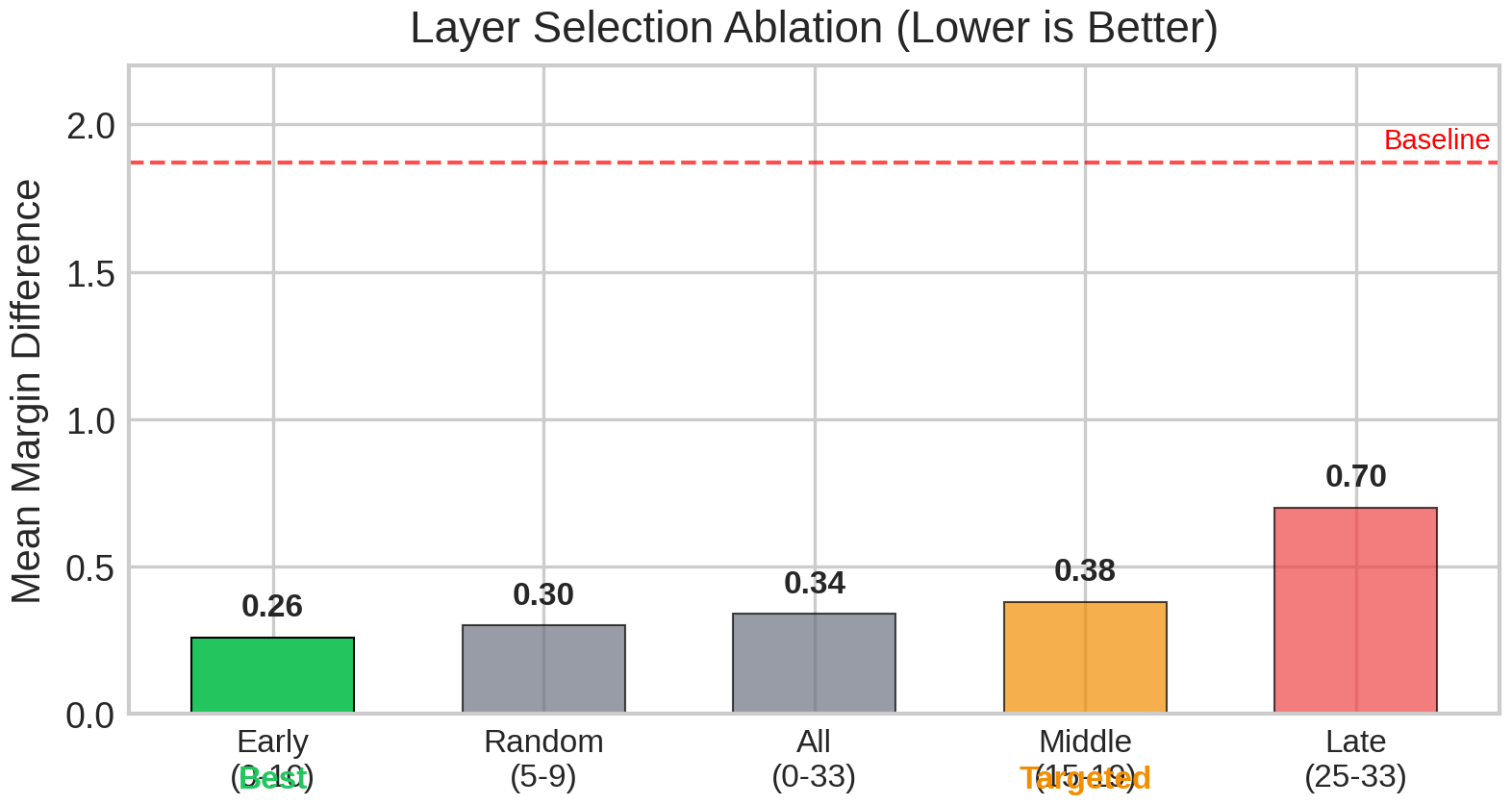}}
\end{figure}

This result has important implications for mechanistic interpretability methodology. Feature 3818 at layer 17 provides a valid \emph{explanation} of where register sensitivity manifests in the model's representations. Our causal validation confirms it influences outputs. However, the optimal \emph{intervention} targets early layers, where representations may be more malleable and where modifications can prevent sensitivity from arising rather than correcting it after manifestation.

We hypothesize that early-layer interventions succeed because they modify representations before register-dependent processing occurs, effectively preventing the sensitivity from developing rather than trying to fix it downstream. This is analogous to preventing an error at its source versus patching it after it has propagated through the system.

\subsection{Training Dynamics}

Figure~\ref{fig:training} shows training dynamics for the combined loss approach on the middle-layer (15 to 19) configuration. Unlike pure consistency training which collapses within the first epoch (accuracy drops to chance level as the model learns to predict all ``Yes''), the combined loss maintains accuracy above 80\% throughout training while steadily reducing margin differences.

\begin{figure*}[t]
\floatconts
  {fig:training}
  {\caption{Training dynamics with combined loss (layers 15 to 19). Consistency loss and mean margin difference drop during training.}}
  {%
  \includegraphics[width=0.49\textwidth]{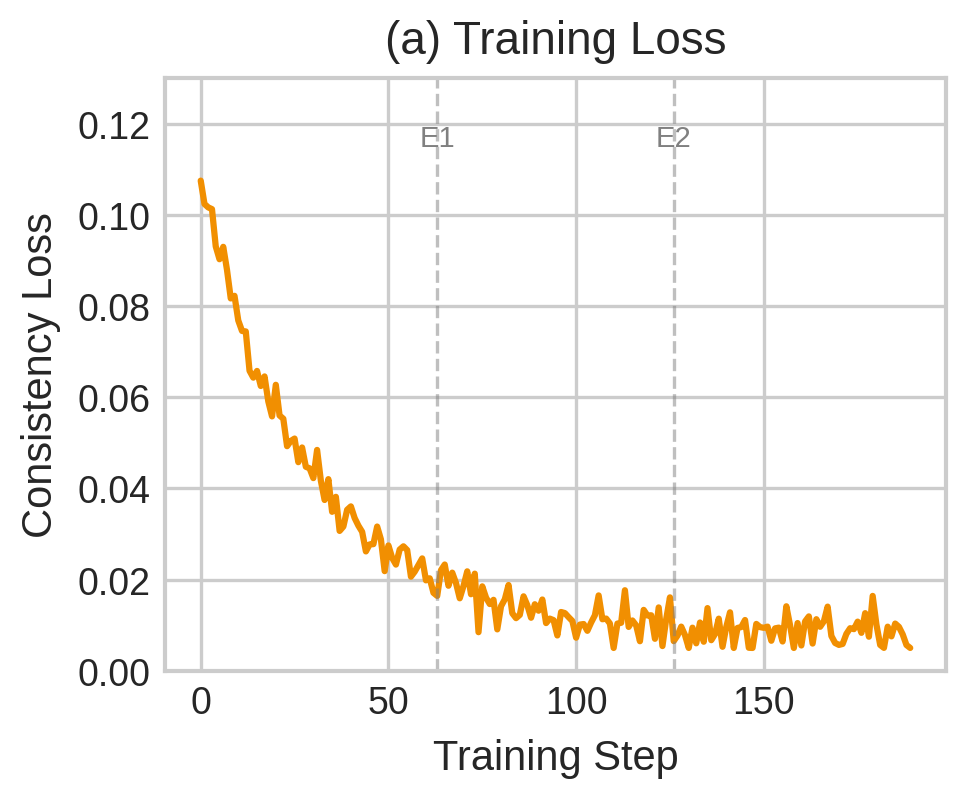}\hfill
  \includegraphics[width=0.49\textwidth]{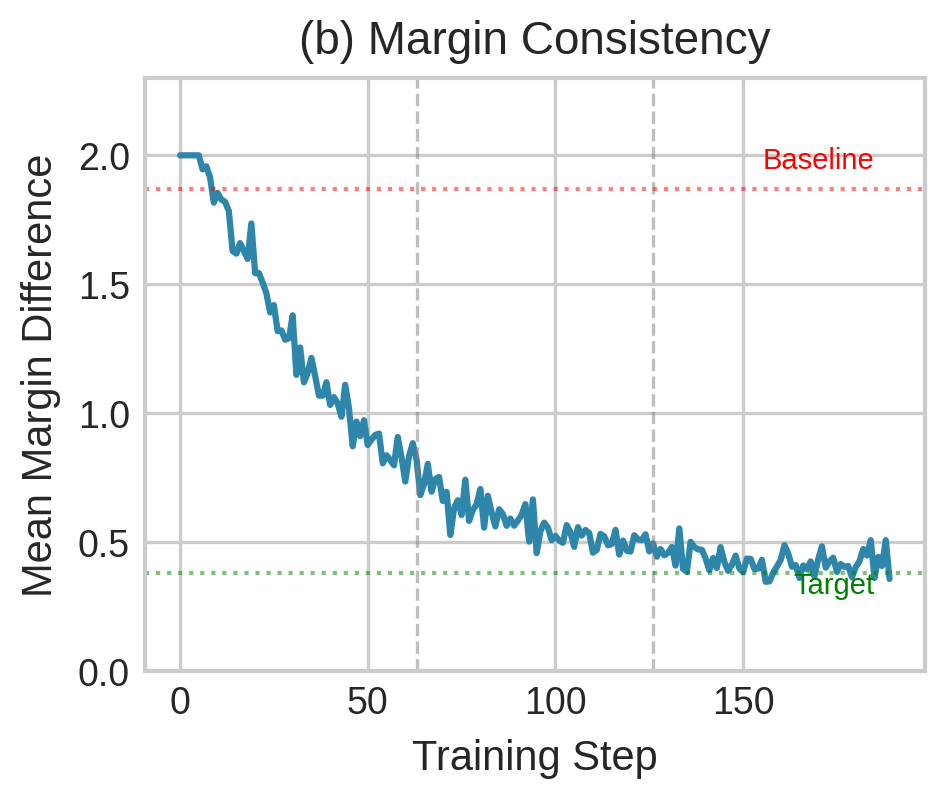}%
  }
\end{figure*}

The training curves show the combined loss balancing both objectives effectively. The consistency loss component decreases rapidly in the first epoch as the model learns to produce similar distributions for paraphrases. Meanwhile, the accuracy component keeps the model grounded, preventing the degenerate solution of predicting the same answer for all questions. By epoch 3, the model achieves low consistency loss (indicating similar predictions for paraphrases) while maintaining high accuracy (indicating correct discriminative behavior). This demonstrates that the combined loss successfully addresses the mode collapse problem inherent in pure self-consistency training.

\subsection{Cross-Dataset Generalization}

To evaluate whether consistency improvements transfer across datasets, we evaluate our LoRA adapters (trained on MIMIC-CXR) on PadChest \citep{bustos2020padchest}, a Spanish chest X-ray dataset with different imaging protocols and patient demographics. We construct a balanced evaluation set of 250 binary presence questions (125 positive, 125 negative) from PadChest to enable meaningful accuracy comparison.

\begin{table}[htbp]
\floatconts
  {tab:cross_dataset}
  {\caption{Cross-dataset generalization to PadChest (balanced $n=250$: 125 yes, 125 no). LoRA training improves both accuracy and consistency on unseen data.}}
  {%
	  \setlength{\tabcolsep}{4pt}
	  \begin{tabular}{@{}lccc@{}}
	  \toprule
	  \bfseries Model & \bfseries Accuracy & \bfseries Margin Diff & \bfseries Flip Rate\\
	  \midrule
	  Baseline & 66.4\% & 1.08 & 13.6\%\\
	  + LoRA & \textbf{69.4\%} & \textbf{0.35} & \textbf{7.8\%}\\
	  \bottomrule
	  \end{tabular}
	  }
	\end{table}
	
	The results show cross-dataset transfer. On PadChest Balanced, flip rate drops from 13.6\% to 7.8\% (42.6\% relative reduction), mean margin difference drops from 1.08 to 0.35 (67.9\% reduction), and accuracy increases from 66.4\% to 69.4\% (+3.0 percentage points), despite training only on MIMIC-CXR.
	
	The flip rate remains non-zero, so these adapters do not fully solve paraphrase sensitivity out of domain. This makes PadChest a useful stress test for future work on calibration, class balance, and distribution shift.

\section{Discussion}
\label{sec:discussion}

Mechanistic interpretability gave us a concrete hypothesis about why paraphrases can change MedGemma's answers: a register-sensitive SAE feature at layer 17 that moves the yes/no margin in some cases. But the layer ablation is a reminder that mechanisms and interventions can live in different places. The feature indicates where the sensitivity shows up, while early-layer LoRA is the most effective fix in our experiments.

\paragraph{Limitations.} We provide a causal patching demonstration for one exemplar flip case, not a full distributional analysis across FlipBank. We also study one model and one task format (binary chest X-ray questions), and our combined loss requires ground truth labels for the accuracy term. PadChest Balanced shows improvement out of domain, but does not eliminate flips.

\paragraph{Clinical implications.} Flip rate alone can miss instability when margins change a lot without crossing zero. For validation, we recommend reporting both flips and margin-based metrics, and checking consistency under common paraphrases before deployment.

\section{Conclusion}
\label{sec:conclusion}

We studied paraphrase sensitivity in MedGemma-4B and found that semantically equivalent questions can change both the answer and the yes/no margin. Using transferred SAEs, we identified a register-sensitive feature at layer 17 and showed with activation patching that it can influence the decision in an exemplar case. We then trained LoRA adapters with a combined consistency and accuracy loss, avoiding mode collapse and improving consistency on MIMIC-CXR and PadChest Balanced. Early-layer adapters performed best in ablation, suggesting that effective interventions can act upstream of where a mechanism appears.

\acks{We thank the creators of Gemma Scope 2 and MedGemma.}

\clearpage
\bibliography{references}

\clearpage
\appendix

\section{Extended Related Work}
\label{apd:related}

\paragraph{Medical Vision-Language Models.} MedGemma \citep{medgemma2025} extends Gemma with medical pre-training. LLaVA-Med \citep{llavamed2024} adapts LLaVA through curriculum learning on medical image-text pairs. LLaVA-Rad \citep{llavarad2024} focuses on radiology with lightweight architectures. RadFM \citep{radfm2023} pre-trains on large-scale radiology data. CheXagent \citep{chen2024chexagent} targets chest X-ray interpretation specifically. These models are evaluated on accuracy using VQA-RAD \citep{lau2018vqarad} and SLAKE \citep{liu2021slake}, but not on consistency under rephrasing.

\paragraph{Consistency and Behavioral Testing.} CheckList \citep{ribeiro2020beyond} introduced systematic behavioral testing under controlled perturbations. \citet{elazar2021measuring} showed that pre-trained models give contradictory answers to logically equivalent questions. Cycle-consistency methods \citep{shah2019cycle} and training approaches \citep{gan2019improving} have been developed to reduce VQA sensitivity to rephrasing.

\paragraph{Mechanistic Interpretability.} Sparse autoencoders decompose activations into interpretable components \citep{cunningham2023sparse, bricken2023monosemanticity}. Activation patching \citep{meng2022locating} tests causal influence. Sparse feature circuits \citep{marks2024sparse} discover causal graphs of information flow.

\section{Metric Definitions}
\label{apd:metrics}

\textbf{Question-level flip rate}: Fraction of questions where at least one paraphrase produces a different binary answer. If a question has 4 paraphrases and 1 flips, the question-level contribution is 100\%.

\textbf{Pair-level flip rate}: Fraction of (original, paraphrase) pairs where binary answers differ. Using the same example, pair-level contribution is 25\%.

\textbf{Mean margin difference}: $\mathbb{E}[|m_{\text{orig}} - m_{\text{para}}|]$ where $m = \log p_{\text{yes}} - \log p_{\text{no}}$. Captures inconsistency even when binary answers match.

\section{Training Hyperparameters}
\label{apd:hyperparams}

\begin{table}[htbp]
\floatconts
  {tab:hyperparams}
  {\caption{Complete hyperparameter specification for combined loss training.}}
  {\begin{tabular}{ll}
  \toprule
  \bfseries Parameter & \bfseries Value\\
  \midrule
  LoRA rank ($r$) & 16\\
  LoRA alpha ($\alpha$) & 32\\
  LoRA dropout & 0.05\\
  Target layers & 15 to 19 (LM only)\\
  Target modules & q,k,v,o; gate,up,down\\
  Learning rate & $2 \times 10^{-4}$\\
  Effective batch size & 8\\
  Warmup steps & 100\\
  Epochs & 3\\
  Optimizer & AdamW\\
  Weight decay & 0.01\\
  Training samples & 500 binary pairs\\
  Accuracy loss weight ($\lambda$) & 1.0\\
  Temperature & 1.0\\
  \bottomrule
  \end{tabular}}
\end{table}

\section{SAE Transfer Validation}
\label{apd:sae}

We validate that Gemma Scope 2 SAEs (trained on base Gemma 2) transfer to MedGemma-4B. This is important because training new SAEs is expensive; reusing existing SAEs enables interpretability research on fine-tuned models.

\begin{table}[htbp]
\floatconts
  {tab:sae_transfer}
  {\caption{SAE transfer validation ($n=100$ prompts per category). Both domains exceed 95\% threshold with $p<0.001$.}}
  {\begin{tabular}{lcc}
  \toprule
  \bfseries Metric & \bfseries Medical & \bfseries General\\
  \midrule
  $R^2$ & $0.9972 \pm 0.0005$ & $0.9974 \pm 0.0006$\\
  FVU & 0.28\% & 0.26\%\\
  L0 (features) & $70 \pm 10$ & $76 \pm 10$\\
  \bottomrule
  \end{tabular}}
\end{table}

Beyond reconstruction, the SAE captures domain-specific features. Feature 203 activates 874 units higher on medical text than general text; Feature 48 shows the opposite pattern (-608 units). This demonstrates semantic structure preservation.

\begin{figure}[htbp]
\floatconts
  {fig:sae_transfer}
  {\caption{SAE transfer: (a) Reconstruction quality, (b) Domain-specific features.}}
  {\includegraphics[width=0.9\linewidth]{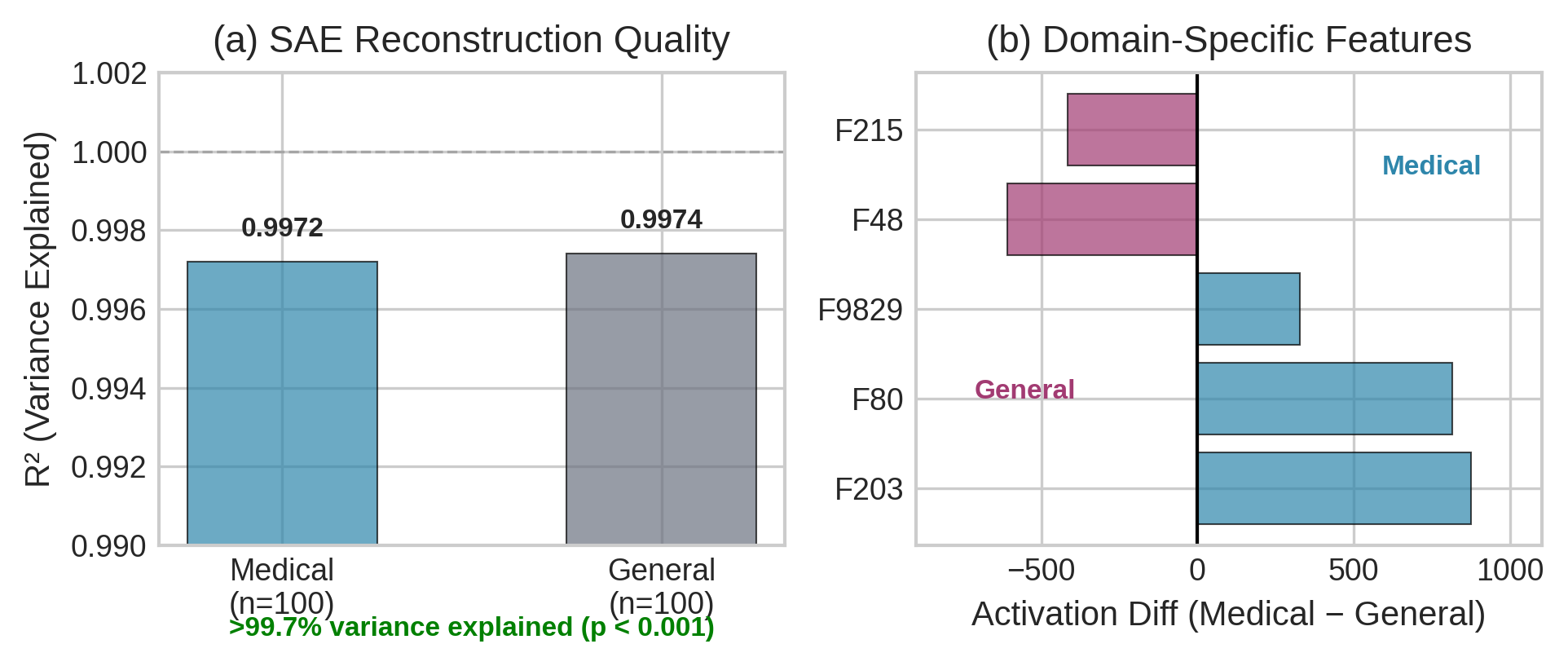}}
\end{figure}

\section{Control Experiment Details}
\label{apd:control}

\begin{table}[htbp]
\floatconts
  {tab:control_features}
  {\caption{Feature specificity ($n=30$ pairs, 300 control measurements). Response rate uses threshold of $|\Delta|>10$ units. Fisher's exact $p = 6.8 \times 10^{-4}$.}}
  {%
  \setlength{\tabcolsep}{3pt}
  \resizebox{\linewidth}{!}{%
  \begin{tabular}{@{}lcc@{}}
  \toprule
  \bfseries Feature & \bfseries Response Rate ($|\Delta|>10$) & \bfseries Mean $|\Delta|$\\
  \midrule
  3818 & 10\% (3/30) & 11.3\\
  Controls (10) & 0\% (0/300) & $<0.5$\\
  \bottomrule
  \end{tabular}%
  }
  }
\end{table}

Feature 3818 is selective: it shows substantial change ($>10$ units) for a small subset of prompt pairs, and the direction can differ across prompts. In this set, deltas exceeding the threshold range from 15 to 185. The 10 control features (matched for baseline activation magnitude) show negligible response across all 300 measurements, with mean $|\Delta| < 0.5$.

\begin{figure}[htbp]
\floatconts
  {fig:control}
  {\caption{Control experiment: Feature 3818 shows substantial changes for a subset of prompt pairs; control features show negligible variation.}}
  {\includegraphics[width=0.9\linewidth]{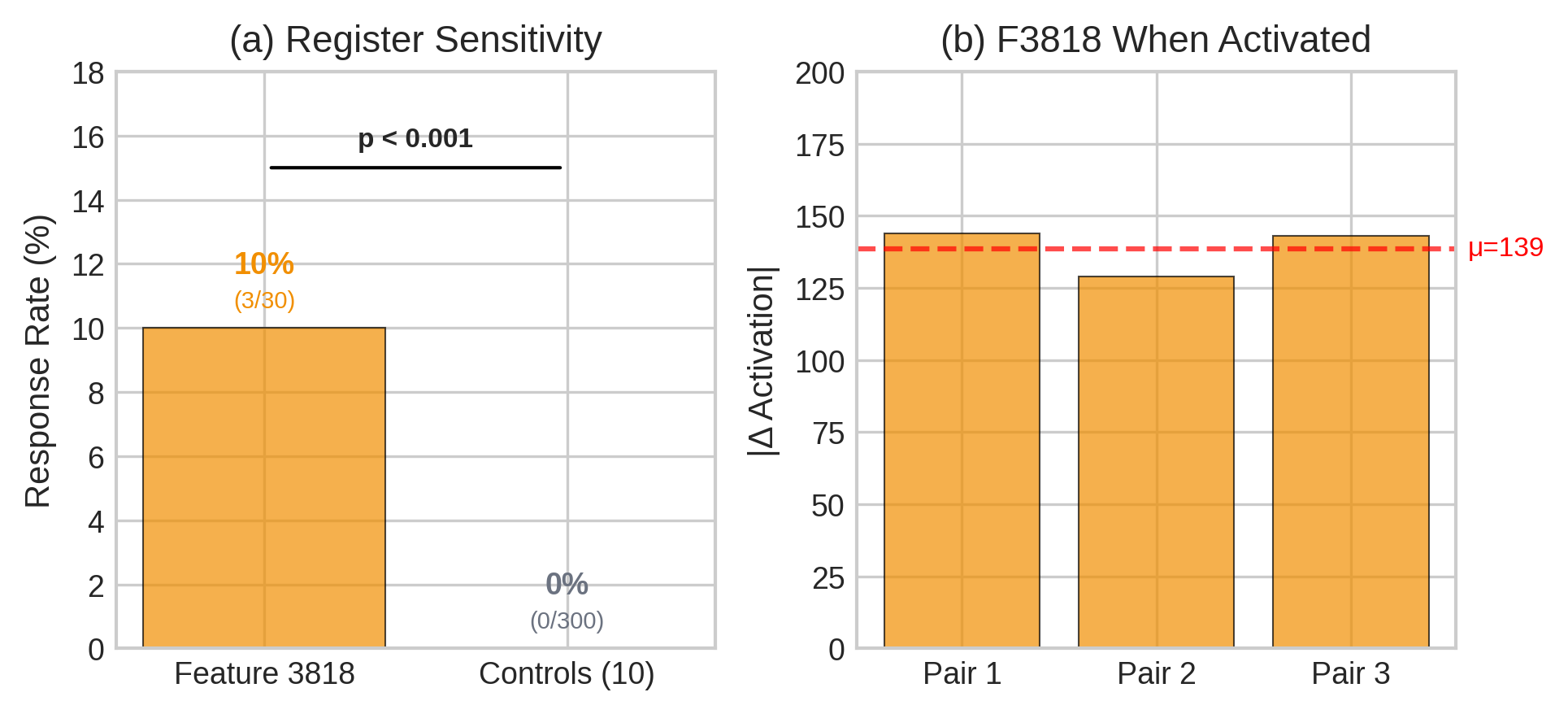}}
\end{figure}

\section{Paraphrase Examples}
\label{apd:examples}

\begin{table}[htbp]
\floatconts
  {tab:para_examples}
  {\caption{Examples from the 30-pair control set where Feature 3818 changes.}}
  {\begin{tabular}{p{3.5cm}cc}
  \toprule
  \bfseries Prompt Pair (Formal / Casual) & \bfseries F3818 & \bfseries F3818\\
  & \bfseries Formal & \bfseries Casual\\
  \midrule
  ``Is there radiographic evidence of pneumo...'' / ``Does the image show air in the chest cav...'' & 144 & 129\\
  \midrule
  ``Is there evidence of pulmonary nodules?'' / ``Can you see any lumps in the lungs?'' & 0 & 185\\
  \midrule
  ``Is there evidence of mass?'' / ``Can you see a tumor?'' & 0 & 140\\
  \bottomrule
  \end{tabular}}
\end{table}

\section{Complete Metric Audit}
\label{apd:audit}

\begin{table}[htbp]
\floatconts
  {tab:metric_audit}
  {\caption{Metric audit on PSF-Med MIMIC-CXR binary questions ($n=158$).}}
  {\begin{tabular}{lcc}
  \toprule
  \bfseries Metric & \bfseries Baseline & \bfseries +LoRA\\
  \midrule
  Binary questions & 158 & 158\\
  Ground truth Yes & 79 & 79\\
  Ground truth No & 79 & 79\\
  \midrule
  Flips & 23 & 7\\
  Flip rate & 14.6\% & 4.4\%\\
  Flip reduction & N/A & 69.6\%\\
  \midrule
  Mean margin diff & 1.63 & 0.33\\
  Margin reduction & N/A & 79.5\%\\
  \midrule
  Correct predictions & 133 & 130\\
  Accuracy & 84.2\% & 82.3\%\\
  \midrule
  Model predicts Yes & N/A & 76\\
  Model predicts No & N/A & 82\\
  \bottomrule
  \end{tabular}}
\end{table}

\paragraph{Statistical tests.} Two-proportion z-test for flip rate: $z=2.87$, $p=0.002$ (one-tailed). Cohen's $h$ effect size: 0.36 (small to medium). Two-proportion z-test for accuracy: $z=0.44$, $p=0.66$ (two-tailed, not significant). Post-hoc power analysis: 85\% power at $\alpha=0.05$.

\section{Mode Collapse in Pure Consistency Training}
\label{apd:mode_collapse}

During our initial experiments, we trained LoRA adapters using only the symmetric KL divergence consistency loss, without any accuracy supervision. The hypothesis was that encouraging the model to produce consistent predictions across paraphrases would naturally lead it to converge on the correct answer. Instead, the model found a degenerate solution that we term mode collapse.

\paragraph{Observed behavior.} After training with pure consistency loss for one epoch, the model learned to predict ``Yes'' for every question regardless of the image content or clinical finding. This trivially minimizes the consistency loss (both paraphrases receive identical ``Yes'' predictions, so the KL divergence is zero) but destroys the model's discriminative ability. Accuracy dropped from 84\% (baseline) to approximately 50\% (the proportion of ground truth ``Yes'' answers in the evaluation set).

\paragraph{Why mode collapse occurs.} Self-consistency objectives without supervision create an underconstrained optimization problem. The model can minimize divergence between paraphrases in two ways: (1) learn to give the same correct answer for both, or (2) learn to give the same arbitrary answer for both. Path (2) is easier because it does not require the model to maintain its understanding of the image or question content. The model learns to ignore the input and produce a constant output distribution.

\paragraph{The combined loss solution.} Adding the cross-entropy accuracy loss provides the missing constraint. The accuracy term penalizes the model for predicting the wrong answer, forcing it to maintain discriminative ability. The consistency term still encourages matching distributions across paraphrases, but now the model must achieve consistency by converging on the correct answer rather than an arbitrary constant.

\paragraph{Empirical validation.} With the combined loss ($\lambda=1.0$), the trained model predicts Yes for 76 questions and No for 82 questions (compared to ground truth of Yes=79, No=79). This balanced prediction distribution confirms that mode collapse has been prevented. The model maintains its ability to distinguish positive from negative cases while achieving improved consistency.

\section{Full Layer Ablation}
\label{apd:ablation}

\begin{table}[htbp]
\floatconts
  {tab:full_ablation}
  {\caption{Complete layer-range ablation on the MIMIC-CXR validation split ($n=355$).}}
  {\begin{tabular}{lcc}
  \toprule
  \bfseries Layers & \bfseries Margin Diff & \bfseries Reduction\\
  \midrule
  Baseline & 1.87 & N/A\\
  \midrule
  Early (0 to 10) & \textbf{0.26} & \textbf{86\%}\\
  Random (5 to 9) & 0.30 & 84\%\\
  All (0 to 33) & 0.34 & 82\%\\
  Middle (15 to 19) & 0.38 & 80\%\\
  Late (25 to 33) & 0.70 & 63\%\\
  \bottomrule
  \end{tabular}}
\end{table}

\section{PadChest Cross-Dataset Evaluation}
\label{apd:padchest}

We evaluate on a balanced subset of 250 binary presence questions from PadChest, converting finding labels to yes/no format. Questions like ``Is there cardiomegaly?'' with answer ``cardiomegaly'' are labeled ``yes'', while ``no\_acute\_abnormality'' is labeled ``no''. The balanced design enables meaningful accuracy comparison.

\begin{table}[htbp]
\floatconts
  {tab:padchest_full}
  {\caption{PadChest Balanced evaluation results ($n=250$).}}
  {%
  \setlength{\tabcolsep}{4pt}
  \begin{tabular}{@{}lcc@{}}
  \toprule
  \bfseries Metric & \bfseries Baseline & \bfseries +LoRA\\
  \midrule
  Flip rate & 13.6\% & \textbf{7.8\%}\\
  Mean margin diff & 1.08 & \textbf{0.35}\\
  Margin reduction & N/A & 67.9\%\\
  Accuracy & 66.4\% & \textbf{69.4\%}\\
  \bottomrule
  \end{tabular}
  }
\end{table}

\paragraph{Key findings.} On PadChest Balanced, the LoRA model reduces flip rate from 13.6\% to 7.8\% (42.6\% relative reduction), reduces mean margin difference from 1.08 to 0.35 (67.9\% reduction), and increases accuracy from 66.4\% to 69.4\% (+3.0 percentage points).

\end{document}